\documentclass{article}
\usepackage{amsmath, amssymb, amsfonts, bbold, dsfont, xfrac, amsthm}
\usepackage{spconf,amsmath,graphicx}
\usepackage{algorithm}
\usepackage{algpseudocode}
\usepackage{amssymb}
\usepackage[utf8]{inputenc}
\usepackage{graphicx}
\usepackage{caption}
 \usepackage{hyperref}
\usepackage{adjustbox}
\usepackage[utf8]{inputenc}
\usepackage{graphicx}
\usepackage{caption}
\newtheorem{theorem}{Theorem}
\newtheorem{lemma}{Lemma}

\newtheorem{corollary}{Corollary}
\newtheorem{assumption}[theorem]{Assumption}

\usepackage{xcolor}

\title{Convergence of Agnostic Federated Averaging\vspace{-5pt}}
\name{Herlock (SeyedAbolfazl) Rahimi and Dionysis Kalogerias\thanks{This work is supported by the NSF under grant CCF 2242215.} \thanks{\textit{Note:} Full proofs to the results presented below are deferred to a forthcoming journal submission.}\vspace{-7pt}}
\address{Department of ECE---Yale University\vspace{-8pt}}
\begin{document}
\ninept
\maketitle
%

% https://cmsworkshops.com/CAMSAP2025/papers/paper_kit.php
\begin{abstract}
\vspace{-5pt}
Federated learning (FL) enables decentralized model training without centralizing raw data. However, practical FL deployments often face a key realistic challenge: Clients participate intermittently in server aggregation and with unknown, possibly biased participation probabilities. Most existing convergence results either assume full-device participation, or rely on knowledge of (in fact uniform) client availability distributions—assumptions that rarely hold in practice. In this work, we characterize the optimization problem that consistently adheres to the stochastic dynamics of the well-known \emph{agnostic Federated Averaging (FedAvg)}  algorithm under random (and variably-sized) client availability, and rigorously establish its convergence for convex, possibly nonsmooth losses, achieving a standard rate of order $\mathcal{O}(1/\sqrt{T})$, where $T$ denotes the aggregation horizon. Our analysis provides the first convergence guarantees for agnostic FedAvg under general, non-uniform, stochastic client participation, without knowledge of the participation distribution. We also empirically demonstrate that agnostic FedAvg in fact outperforms common (and suboptimal) weighted aggregation FedAvg variants, even with server-side knowledge of participation weights. 
\end{abstract}
\vspace{-4pt}
\vspace{-9pt}
\section{Introduction}
\vspace{-7pt}
\label{sec:intro}

Federated Learning (FL) is an established decentralized machine learning paradigm in which clients collaboratively train a global model without sharing their raw data, thereby preserving privacy and ensuring compliance with data protection regulations~\cite{mcmahan2017communication, kairouz2021advances}. Each client performs local updates on its private dataset and transmits model parameters or gradients to a central server for aggregation. This setup has proven especially useful in privacy-sensitive domains such as mobile health, financial services, and personalized recommendation systems, where raw data cannot be centralized~\cite{hard2018federated, ramaswamy2019federated, yang2019federated, Melis2019}.

While FL is appealing in theory, its practical deployment faces several challenges. Real-world clients often suffer from intermittent connectivity, limited computing power, and statistically heterogeneous data~\cite{Geyer2017, konecny2016federated, wang2021field, nishio2019client}. As a result, most FL systems rely on \emph{partial participation}, where only a randomly selected subset of clients is active during each round of training~\cite{yang2019federated, bonawitz2019towards, hard2020training, chen2020optimal}. In this regime, the central server aggregates updates only from available clients and broadcasts the averaged model back to all participants. Partial participation has become a \textit{de facto} design choice in modern FL infrastructure~\cite{bonawitz2019towards, hard2020training, nishio2019client}.

The classical \textit{Federated Averaging} algorithm (\textit{FedAvg}) \cite{mcmahan2017communication} tacitly assumes either full participation or uniform sampling of clients at each communication round. However, in realistic settings, client availability is governed by complex and often unknown (even to the server) dynamics—such as battery level, device usage patterns, and network service conditions~\cite{li2020federated, bonawitz2019towards, nishio2019client, eichner2019semi}. Consequently, the participation frequency of each client varies, leading to a fundamentally biased sampling process\footnote{Hereafter, we interchangeably use the terms ``client participation" and ``server sampling" to address the same issue.}. Yet, this process is typically not observed by the server. This raises the central questions motivating this work:

\begin{quote}
\centering
\emph{\textbf{Does (agnostic) FedAvg converge under possibly unknown, non-uniform client participation?\\ If so, what objective does it optimize?}}
\end{quote}

To explore these questions, let us formalize a supervised learning setting. Let \( \mathfrak{X} \subset \mathbb{R}^d \) denote the input space and \( \mathcal{C} = \{1, \dots, C\} \) the label space\footnote{We consider classification without loss of generality.}. Each client \( i \in [N] \) holds a local dataset \( D_i = \{(X_i^j, Y_i^j)\}_{j=1}^{n_i} \), drawn from the empirical distribution \( \mathcal{D}_i \), and optimizes its local loss expressed by
\begin{align}
\label{equation:local-objective}
    f(\theta; D_i) = \frac{1}{n_i} \sum_{j=1}^{n_i} \ell(m(X_i^j, \theta), Y_i^j),
\end{align}
where \( \theta \in \Theta \subseteq \mathbb{R}^{d'} \) is the parameter (vector) of a common model \( m: \mathfrak{X} \times \Theta \to \mathcal{C} \), and \( \ell: \mathcal{C} \times \mathcal{C} \to \mathbb{R}_+ \) is a standard loss function (e.g., cross-entropy). Users typically optimize their local parameters via stochastic gradient methods (SGD) and then \textit{attempt} to transmit the local parameters to the server~\cite{mcmahan2017communication, konecny2016federated}.

While previous studies assume artificial cases for the participation processes in which either all or a fixed number of users transmit their parameters to the server  (chosen with known or estimable probabilities) ~\cite{nishio2019client, cho2020client, chen2020optimal, reisizadeh2020fedpaq}, herein we consider a general, most versatile participation process, where at each communication round, there is a probability $q(\mathcal{A})\geq 0$ that the user subset $\mathcal{A}\subseteq[N]$ can transmit their parameters to the server. Moreover, we posit no further restrictions over $q$, e.g.,  its estimation or access by the server~\cite{haddadpour2021federated, dio2024restricted}.

\begin{algorithm}[t]
\caption{Agnostic Federated Averaging (FedAvg)}
\label{alg:agnostic-fedavg}
\begin{algorithmic}[1]
\State \textbf{Initialize:} \( \theta_i^{0} = \mathbf{0} \), \( i \in [N] \), $S$, $H$, $\eta_\theta>0$, $\mathcal{C}\subset \Theta$.
\For{\( t= 1,2, \dots, TH \)}
    \If{\( t \mod H = 0 \)} \Comment{\textbf{\textit{Global Communication}}}
        \State Users transmit local parameters to server: \( S^t \subseteq [N] \).
        \State Server \textbf{aggregates}: \( \hat{\theta}^t = \frac{1}{|S^t|} \sum_{i \in S^t} \theta_i^{t-1} \).
        \State Server \textbf{broadcasts} \( \hat{\theta}^t \):
        \( \theta_i^t = \hat{\theta}^t, \, \forall i \in [N] \)
    \Else \Comment{\textbf{\textit{Local Update}}}
        \State Users tune their local models with SGD ($H$ rounds):
        \State \( \theta_i^t = \Pi_\mathcal{C}\left( \theta_i^{t-1} - \eta_\theta \nabla_\theta f_i(\theta_i^{t-1}; \xi_i^t) \right), \, \forall i \in [N] \) 
    \EndIf
\EndFor
\State \textbf{Return:} \( \frac{1}{S} \sum_{\tau=1}^{S} \hat{\theta}^{\tau H} \)
\end{algorithmic}
\end{algorithm}

A widely used (and straight-forward) \textit{agnostic} FedAvg variant to tackle such a problem is outlined in Algorithm \ref{alg:agnostic-fedavg} (see Section \ref{sec:convergence} for a complete technical description), where all clients locally optimize their local objective for $H$ consecutive rounds and then (\textit{if} available at time $t$) transmit their updated parameter to the server. The server aggregates the received parameters via an \textit{simple unweighted mean} (because of lack of knowledge over the $q$'s) and broadcasts the results to all users~\cite{nishio2019client, li2020federated}.

Despite its empirical success, whether (agnostic) FedAvg converges under general, unknown participation processes remains an open theoretical question. To the best of our knowledge, convergence of FedAvg has not been established even under the more restrictive setting where a fixed number of clients participate in each round and their sampling probabilities are known~\cite{reisizadeh2020fedpaq, haddadpour2021federated, li2020federated}. This motivates a deeper investigation into the algorithm’s behavior under agnostic participation regimes.

Recent works have attempted to account for client heterogeneity in both data and participation through a variety of methods, including improved learning algorithms~\cite{karimireddy2020scaffold, reddi2021adaptive}, control variates~\cite{li2020federated}, and client selection strategies~\cite{chen2020optimal, zhao2018federated, nishio2019client}. These approaches typically assume knowledge of the client participation probabilities, or rely on auxiliary ---usually statistically expensive--- estimation mechanisms to approximate them~\cite{nishio2019client, cho2020client}. To date, no existing work has rigorously studied the convergence behavior of FedAvg under an \emph{entirely unknown and freely highly non-uniform} client sampling distribution, making this an open and pressing theoretical gap.

\textbf{\textit{Our work fills this theoretical gap:}} We identify the objective that is naturally implied by agnostic FedAvg, and (for the first time) establish convergence of agnostic FedAvg within the tractable class of convex but possibly nonsmooth loss functions under such an objective. Our main technical innovation lies in formalizing the client sampling process via a generic probabilistic model over user subsets (otherwise completely unknown to the server), and showing that simple uniform aggregation of available parameters (i.e., line $5$ of Algorithm \ref{alg:agnostic-fedavg}) in fact minimizes a well-defined objective function \textit{canonically induced by} the aforementioned probabilistic model.

Our contributions may be itemized as follows:
\vspace{-3pt}
\begin{itemize}
    \item We propose a rigorous and natural generic stochastic model of client availability that induces non-uniform but hidden sampling probabilities, reflecting real FL system behavior.
    \vspace{-2pt}
    \item We discover and introduce the global optimization problem that FedAvg solves without knowledge of the client sampling probabilities (i.e., aggregating users via simple averaging).
    \vspace{-2pt}
    \item We provide the first full-fledged convergence analysis for FedAvg in this setting, establishing a rate of order \( \mathcal{O}(1/\sqrt{T}) \) for convex objectives under standard conditions, where $T$ denotes the aggregation horizon. This analysis can be applied to settings with fixed-sized user participation as well~\cite{Nguyen2022, Reddi2016,shamir2013}.
\end{itemize}

\vspace{-11pt}
\section{Discovering the Problem Setup}
\vspace{-8pt}
\label{sec:problem-setup}
We model a realistic FL system in which only a subset of clients is available for communication at any given round. Let there be \( N \) clients indexed by \( [N] = \{1, \dots, N\} \). At each global round \( t \), a random subset \( S^t \subseteq [N] \) of clients becomes available to transmit their updated models to the server. This stochastic availability arises due to different phenomena like hardware, network, and behavioral constraints, as documented in practical FL deployments~\cite{li2020federated, yang2019federated}.
At each communication round, there are $2^N$ possibilities for the subset of users available to the server. We denote the possible subsets by $\mathcal{A}_1, \ldots, \mathcal{A}_{2^N}$ with respective probabilities $q(\mathcal{A}_1),\ldots, q(\mathcal{A}_{2^N})$. Now, consider the marginal weight distribution defined as\footnote{with the extra assumption that $q(\emptyset)=0$, and the convention $\frac{0}{0}=0$.}:
\begin{equation}
\label{eq:availability}
   \boxed{
    p_i = \sum_{j=1}^{2^N} \frac{q(\mathcal{A}_j)}{|\mathcal{A}_j|} \mathbb{I}[i \in \mathcal{A}_j], \quad i\in[N],
    }
\end{equation}
which denotes an adjusted (by the size of each subset) \textit{availability} or \textit{survival} of each user $i$. The next lemma proves that the $p_i$'s form a \textit{probability} distribution over users.
\vspace{-3pt}
\begin{lemma}
The marginal weights \( \{p_i\}_{i=1}^N \) form a valid probability distribution over clients.
\end{lemma}
\vspace{-10.5pt}
\begin{proof}
Non-negativity is clear. To show normalization, we can write
\begin{align*}
\sum_{i=1}^N p_i &=  \sum_{j=1}^{2^N}
\frac{q(\mathcal{A}_j)}{|\mathcal{A}_j|} \sum_{i=1}^N\mathbb{I}[i \in \mathcal{A}_j] 
%\\&= \sum_{j=1}^{2^N} \frac{q(\mathcal{A}_j)}{|\mathcal{A}_j|} \sum_{i \in \mathcal{A}_j} 1 
= \sum_{j=1}^{2^N} q(\mathcal{A}_j) = 1,
\end{align*}
completing the proof.
\end{proof}
\vspace{-5pt}

Utilizing the client survival distribution induced by the probabilities $\{ p_i\}_i$, we now introduce the global optimization problem
\begin{align}
\label{equation:global-objective}
\boxed{
    \inf_{\theta \in \Theta} \sum_{i=1}^{N} p_i f(\theta; D_i),
    }
\end{align}
and \textit{conjecture} that it is in fact this problem that is solved by (agnostic) FedAvg (see Algorithm \ref{alg:agnostic-fedavg}). 

To see why this could very well be case, we note that  Algorithm \ref{alg:agnostic-fedavg} is fundamentally a stochastic approximation scheme. Therefore, a potential objective that may be optimized by such a scheme must reflect the statistical dynamics of that scheme. In fact, line $5$ of Algorithm \ref{alg:agnostic-fedavg} indeed implies a uniform distribution but over the \textit{available users}, that is, \textit{after} the user survival subset $S^t$ has been realized. This fact  naturally motivates a hierarchical probabilistic experiment: \textit{First}, a random subset $S^t$ is selected, and \textit{then} client $i$ is chosen from $S^t$ uniformly at random (provided of course that $i\in S^t$; otherwise this event has zero conditional probability). In other words, we have the Bayesian interpretation
\begin{equation}
p_i = \sum_{j}
\underbrace{
      P(\text{user }i\text{ is selected}|S^t=\mathcal{A}_j)}_{=\frac{1}{{|\mathcal{A}_j|} }\mathbb{I}[i \in \mathcal{A}_j]} 
    \underbrace{P(S^t=\mathcal{A}_j)}_{=q(\mathcal{A}_j)},
\end{equation}
which indeed implies uniform user selection (being consistent with line $5$ of Algorithm \ref{alg:agnostic-fedavg}), however subject to the user's chances of availability (i.e., the chances that user belongs to the random subset $S^t$).

%As in all other areas of ML, the weight of each sample in the objective reflects its sampling probability, $1/{|A|}$ in a dataset of size $|A|$, here the weight of each user intuitionally should reflect the sampling marginal of that specific user as well. To calculate how often a user is available to the server, one could see, normalizing the probabilities that user is available to the server, to the size of the subset selected, forms a natural distribution over users. 

The crucial insight herein is that user importance in the agnostic FL setting (and as weighted through the $p_i$'s in problem \eqref{equation:global-objective}) is not dictated by data relevance or label diversity, but by the statistical rate of user availability/survival ---a side effect of the communication process (whether the server knows/controls it \textit{or not}), which is embedded in agnostic FedAvg by construction. While it has long been the case that the FL objective is chosen regardless of the availability of users (even with the aggregation step later modified to somehow adapt to the objective ---see also Section \ref{sec:experiments}), our discussion above indicates that the objective itself is induced by the user selection dynamics. Choosing the distribution of the $p_i$'s arbitrarily and then trying to explain convergence of agnostic FedAvg is an ill-posed problem by construction, since such a choice imposes artificial (or ``on demand") constraints on user availability (such a situation may be tackled by using robustification techniques, which are outside of the scope of standard/vanilla FL). 

%\textbf{Starting from ps, and then trying to explain convergence of agnostic FedAvg under this choice is an ill posed problem by construction, because that put constraints on the availability, and not the other way around }

%\dk{Motivation/Explanation} The objective introduced in problem \eqref{equation:global-objective} is naturally motivated  

%\dk{Use conditional explanation: probability of selecting $i$ uniformly, given that....}

%Participation process is indeed reflected in the operation of the FedAvg algorithm

%we would later in \ref{sec:convergence}, as one of the main contributions of our work, prove that the \ref{alg:agnostic-fedavg} optimizes this objective and we would provide the convergence analysis for the convex case.  
 
 The discussion above is reinforced by the fact that existing convergence analysis of (agnostic) FedAvg in the literature assumes (to the best of knowledge) that the weights \( \{p_i\} \) are either known, also assuming full client participation, or uniform~\cite{mcmahan2017communication, chen2020optimal} (i.e., $p_i=1/N$) while allowing fixed-size partial participation, the latter being clearly a special corner case of the statistical model outlined above. Lastly, the proposed model subsumes various other availability models considered in the literature as well; see, e.g., those in \cite{dio2024restricted} and \cite{dal2023federated}.

 %Moreover, while other studies start from $p_i$ and prove the convergence only in case of full device participation, or uniform same-frequent participation in partial case, we have found this is not the case in general.

\section{Convergence Analysis of Agnostic FedAvg}
\vspace{-6pt}
\label{sec:convergence}
We develop a convergence analysis of agnostic FedAvg under standard assumptions. The algorithm incorporates two sources of randomness: First, for each user \( i \in [N] \), the random element \( \xi_i \) denotes a mini-batch of size $b$ sampled uniformly without replacement from their local dataset. At time \( t \), we denote this by \( \xi_i^t \). Second, let \( S^t \subseteq [N] \) denote the subset of users selected (\textit{iid}) at round \( t \). This selection is based on the distribution induced by the $q$'s already covered. Within either a global or local round in the operation of Algorithm \ref{alg:agnostic-fedavg}, we consider both sources of randomness with the understanding that during global rounds the sampled (outputted) variables $\xi_i^t$ are not used, and during local rounds the outcome $S^t$ is not used.

That said, we define the natural filtration generated by the agnostic FedAvg process as $\{\mathcal{F}_t\}_t$, with each $\sigma$-algebra defined as
\[
\mathcal{F}_t := \sigma\left( \theta_i^s, \xi_i^s, S^s : s \leq t,\, i \in [N] \right),
\]
capturing the information generated by model states, mini-batch selections, and user participation histories up to round \( t \). The iterates \( \theta_i^t \) evolve as a Markov process with respect to this filtration, depending only on \( \theta_i^{t-1} \), \( \xi_i^{t-1} \), and \( S^{t-1} \) and the fresh \(\xi_i^t\) or \(\ S^t\). Our assumptions on the problem class are as follows.
\vspace{-3pt}
\begin{assumption}[\textbf{Convexity}]
\label{assump:convex}
The loss functions  
%$\ell(\cdot)$, and hence, 
$f_i(\cdot):=f(\cdot; D_i), i\in [N]$  are all convex on $\Theta$.
\end{assumption}
\vspace{-9pt}
\begin{assumption}[\textbf{Bounded Local Gradient Variance}]
\label{assump:variance}
For \( i \in [N] \), it is true that
\[
\sup_{\theta \in \Theta}\mathbb{E}_{\xi_i} \left[ \left\| \nabla f(\theta; \xi_i) - \nabla f_i(\theta) \right\|^2 \right] \leq \sigma_i^2.
\]
We also define the global variance upper bound \( \sigma^2 := \sum_{k=1}^N p_i \sigma_i^2 \).
\end{assumption}
\vspace{-10pt}
\begin{assumption}[\textbf{Bounded Gradient Norm}]
\label{assump:gradient_bound}
For \( i \in [N] \), it is also true that
\[
\sup_{\theta \in \Theta}\mathbb{E}_{\xi_i} \left[ \left\| \nabla f(\theta; \xi_i) \right\|^2 \right] \leq G^2.
\]
\end{assumption}

We further tacitly assume that $\mathcal{C}$ is convex compact with the Euclidean projection onto $\mathcal{C}$ denoted as $\Pi_\mathcal{C}(\cdot)$, as well as that an optimal solution $\theta^* \in\mathcal{C}$ to problem \eqref{equation:global-objective} exists. Since each (convex) $f_i(\cdot)$ is also locally Lipchitz and $\mathcal{C}$ is compact, $f_i(\cdot)$ is Lipchitz on $\mathcal{C}$ as well, say with a common constant $\ell$. Under these circumstances, we proceed to establish convergence of Algorithm \ref{alg:agnostic-fedavg}.

\vspace{-4pt}
\subsection{Preliminary Lemmata}

We first analyze the standard effect of a single projected stochastic gradient step with mini-batch noise at the client side. 
\vspace{-4pt}
\begin{lemma}[\textbf{One-Step Progress}]
\label{lemma:onestep}
Let \( \theta_i^{t} = \Pi_{\mathcal{C}}(\theta_i^{t-1} - \eta g_i^{t}) \), $i\in[N]$, where \( g_i^{t} \) is any unbiased (sub)gradient estimator, i.e. (cf. line $9$ of Algorithm \ref{alg:agnostic-fedavg}),
$\mathbb{E}[g_i^t \mid \mathcal{F}_{t-1}] = \nabla f_i(\theta_i^{t-1})$.
Let Assumptions~\ref{assump:variance} and~\ref{assump:gradient_bound} be in effect. At local rounds, it is true that 
\begin{align}
\mathbb{E} \left[ \| \theta_i^t - \theta^* \|^2 \mid \mathcal{F}_{t-1} \right] 
&\leq \| \theta_i^{t-1} - \theta^* \|^2 \nonumber \\
&\quad - 2\eta \left(f_i(\theta_i^{t-1}) - f_i(\theta^*) \right) \nonumber \\
&\quad + 2\eta^2 \sigma^2 + 2\eta^2 G^2. \tag{l1-(1)} \label{eq:l1}
\end{align}
\end{lemma}

% \textit{Proof Sketch.} \dk{Maybe remove sketches? Discuss} The result follows from expanding the squared norm, applying non-expansiveness of the projection operator, using unbiasedness of gradients, and bounding the variance and gradient norm. See the journal version for the full derivation.

Next, we bound the distance between the parameter vectors of two arbitrary users between two local communication rounds.

\vspace{-4pt}
\label{lemma:local_divergence}
\begin{lemma}[\textbf{Local Parameter Divergence}]
Let \( i, j \in [N] \) be two users, and let \( \tau_i, \tau_j \in [SH, SH +H] \) denote any local steps occurring between two consecutive communication rounds \( S \) and \( S+1 \). Under Assumption~\ref{assump:gradient_bound}, we have
\[
\mathbb{E} \left[\|\theta_i^{\tau_i} - \theta_j^{\tau_j}\| \big| \mathcal{F}_{SH}\right] \leq 4\eta G H.
\]
\end{lemma}

%\textit{Proof Sketch.} The result follows by telescoping the difference between \( \theta_i^{\tau_i} \) and \( \theta_j^{\tau_j} \), applying triangle inequality and using the boundedness of the gradient norm across \( H \) steps. Details are deferred to the journal version.

Lipschitz continuity of the $f_i$'s easily implies the following fact.\vspace{-2pt}
\begin{corollary}[\textbf{Local Value Divergence}]
\label{corollary:lipschitz_difference}
For users \( i, j \in [N] \) and corresponding local steps \( \tau_i, \tau_j \in [SH, SH+H] \), it is true that
\[
\mathbb{E}\left[ |f_i(\theta_i^{\tau_i}) - f_i(\theta_j^{\tau_j})| \big| \mathcal{F}_{SH}\right] \leq 2\ell \eta G H. \tag{C1-(11)} \label{C1-1}
\]
\end{corollary}

%\textit{Proof.} Follows directly from the \(\ell\)-Lipschitz continuity of \(f_i\) and Lemma~\ref{lemma:local_divergence}, noting that
%\[
%|f_i(\theta) - f_i(\theta')| \leq \ell \|\theta - \theta'\|.
%\]

\subsection{Client Availability and Global Model Update}

The aggregation step at the server at global round time \( t \) is given by
\[
\hat{\theta}^{t} = \frac{1}{|S^t|} \sum_{j \in S^t} \theta_j^{t-1}.
\]
%\dk{need to explain this in detail, as it is the most novel part of the proof}
%The expected distance of the global iterate to the optimum can be bounded by:
A main result of central relevance in this work explicitly relates server aggregation with the survival probabilities $\{p_i\}_i$.
\vspace{-2pt}
\begin{lemma}[\textbf{Sample-to-Model Inequality}]
    At every global round $t$, it is true that
\begin{equation}\label{GlobalToLocal} 
    \mathbb{E} \big[\|\hat{\theta}^{t} - \theta^*\|^2\mid \mathcal{F}_{t-1}\big]
    \le
    \sum_{i \in [N]} p_i \cdot  \|\theta_i^{t-1}  - \theta^*\|^2.
\end{equation}
\end{lemma}
\vspace{-12pt}
\begin{proof}
First, Jensen implies that
\begin{align}\nonumber
        \mathbb{E} \big[\|\hat{\theta}^{t} - \theta^*\|^2\mid \mathcal{F}_{t-1}\big] =  \mathbb{E}_{S^{t}}\Bigg[ \Bigg\| \frac{1}{|S^t|} \sum_{i \in S^{t}} \theta_i^{t-1} - \theta^* \Bigg\|^2  \Bigg| \mathcal{F}_{t-1}\Bigg] \\
\leq  \mathbb{E}_{S^{t}} \Bigg[ \frac{1}{|S^t|} \sum_{i \in S^{t}} \|\theta_i^{t-1} - \theta^*\|^2 \Bigg | \mathcal{F}_{t-1}\Bigg]. \nonumber
    \end{align}
Expanding the expectation on the right-hand side, we have
\begin{align}\nonumber
&\hspace{-8pt}\mathbb{E} \big[\|\hat{\theta}^{t} - \theta^*\|^2\mid \mathcal{F}_{t-1}\big] \\ \nonumber &
\leq  \sum_{\mathcal{A}} \left[{P}[S^{t} = \mathcal{A}] \cdot \frac{1}{|\mathcal{A}|} \sum_{i \in \mathcal{A}} \|\theta_i^{t-1} - \theta^*\|^2\right] \\ \nonumber &
 = \sum_{\mathcal{A}} \Bigg[{P}[S^{t}=\mathcal{A}] \cdot \frac{1}{|\mathcal{A}|} \sum_{i \in [N]} \mathbb{I}[i\in \mathcal{A}] \|\theta_i^{t-1} - \theta^*\|^2\Bigg]\\ \nonumber &
=  \sum_{i \in [N]} \Bigg[\sum_{\mathcal{A}}\frac{1}{|\mathcal{A}|}{P}[S_{t}=\mathcal{A}]\mathbb{I}[i \in \mathcal{A}]\Bigg] \cdot \|\theta_i^{t-1} - \theta^*\|^2\\ \nonumber
&
=  \sum_{i \in [N]} p_i \cdot  \|\theta_i^{t-1}  - \theta^*\|^2, 
\end{align}
and we are done.
\end{proof}
%proof. The result follows after applying the Jensen inequality and the definition of $p_i$'s. 
\vspace{-12pt}
\subsection{Recursive Descent and Putting It Altogether}

Applying inequality~\eqref{GlobalToLocal} to the terminal round \( t = TH \) and taking expectation on both sides, we first obtain
\begin{equation}\label{GTL2}
\mathbb{E} \big[\|\hat{\theta}^{TH} - \theta^*\|^2 \big] \leq \sum_{i\in[N]} p_i \mathbb{E} \big[\| \theta_i^{TH-1} - \theta^* \|^2 \big].
\end{equation}
Now, invoking Lemma~\ref{lemma:onestep} and~\ref{lemma:local_divergence}, we can relate \( \theta_i^{TH-1} \) to \( \hat{\theta}^{TH} \) through recursive analysis. Specifically, for each user \( i \), we apply Lemma~\ref{lemma:onestep} across \( H \) steps and use the Lipschitz continuity of \( f_i \), together with the coupling bound in Lemma~\ref{lemma:local_divergence}, yielding
\begin{align}
\mathbb{E} \big[ \| \theta_i^{TH-1} - \theta^* \|^2 \big] 
&\leq \mathbb{E} \big[ \| \theta_i^{(T-1)H} - \theta^* \|^2 \big] \nonumber \\
&\quad - 2\eta H \mathbb{E} \big[f_i(\hat{\theta}^{TH}) - f_i(\theta^*) \big] \nonumber \\
&\quad + H\left(2\eta^2 \sigma^2 + 3\eta^2 G^2 + 8\ell \eta^2 G H\right).
\end{align}
Combining with \eqref{GTL2} and noting that $\theta_i^{(T-1)H}=\hat{\theta}^{(T-1)H}$ , we get
\vspace{-2pt}
\begin{multline}
\mathbb{E} \big[ \| \hat{\theta}^{TH} - \theta^* \|^2 \big] 
\leq \mathbb{E} \big[ \| \hat{\theta}^{(T-1)H} - \theta^* \|^2 \big] \\
 - 2\eta H \mathbb{E} \big[f(\hat{\theta}^{TH}) - f(\theta^*) \big] \\
+ H \left(2\eta^2 \sigma^2 + 3\eta^2 G^2 + 8\ell \eta^2 G H \right).
\end{multline}
Rearranging, averaging over global rounds \( \tau = H, 2H, \ldots, TH \) and calling Jensen once more, we arrive at the rate expression
%\begin{align}
%\mathbb{E} \left[ f(\hat{\theta}^{TH}) - f(\theta^*) \right]
%&\leq \frac{\mathbb{E} \left[ \| \hat{\theta}^{(T-1)H} - \theta^* \|^2 - \| \hat{\theta}^{TH} - \theta^* \|^2 \right]}{2\eta H} \nonumber \\
%&\quad + 1.5 \eta \sigma^2 + 3 \eta G^2 + 4 \ell \eta G H.
%\end{align}
%\subsection{Final Convergence Rate}
\begin{multline}
\mathbb{E} \left[ f\left( \frac{1}{T} \sum_{\tau = 1}^T \hat{\theta}_{\tau H} \right) - f(\theta^*) \right]
\\\leq \frac{\|\hat{\theta}_0 - \theta^*\|^2}{2\eta H T} +  \eta \sigma^2 + 1.5 \eta G^2 + 4 \ell \eta G H,
\end{multline}
finally leading to our main result.
%where \( D := \|\hat{\theta}_0 - \theta^*\|^2 \) and 
%choosing \( \eta = \frac{c}{\sqrt{T H}} \), we get the final convergence bound:
%\begin{multline}
%\mathbb{E} \left[ f\left( \frac{1}{T} \sum_{\tau = 1}^T \hat{\theta}_{\tau H} \right) - f(\theta^*) \right]
%\leq\\ \left( \frac{D}{2c} + 1.5c \sigma^2 + 3c G^2 + 4 \ell c G H \right) \cdot \frac{1}{\sqrt{T H}}.
%\end{multline}
\vspace{-4pt}
\begin{theorem}[\textbf{Convergence of Agnostic FedAvg}]
\label{thm:main}
Let Assumptions \ref{assump:convex}, \ref{assump:variance} and \ref{assump:gradient_bound}  be in effect. Then, (projected) agnostic FedAvg satisfies
\[
\boxed{
\mathbb{E}\left[f\left( \frac{1}{T} \sum_{t=1}^T \hat{\theta}_{tH} \right) - f(\theta^*)\right] = \mathcal{O}\left(\frac{1}{\sqrt{T}}\right),
}
\]
with appropriate step size \( \eta = \Theta\big(1/\sqrt{T H}\big) \).
\end{theorem}
%\textit{Note.} All detailed derivations, including bounds involving correlations in user selection, generalization to non-i.i.d. data distributions, and tightness analysis, are deferred to the extended journal version.
\vspace{-14pt}

\section{Experimental Evaluation}
\vspace{-7pt}
\label{sec:experiments}

While the convergence of FedAvg under full-device participation has long been established~\cite{zhao2018federated}, our experiments below reveal that the convergence behavior of commonly employed \emph{weighted FedAvg} under partial participation—particularly with imbalanced user selection probabilities—is clearly suboptimal (relative to the objective of problem \eqref{equation:global-objective}). Indeed, prior work \cite{zhao2018federated} has considered a special case of user sampling where, at each round, a fixed-size subset $|S^t| = M$ is selected, and the selection probabilities $p_i$ are known. In this case, weighted FedAvg aggregates clients as
\begin{equation}
    \label{equation:other algorithms}
    \hat{\theta}^t = \frac{N}{|S^t|} \sum_{i \in S^t} p_i \theta_i^{t-1}.
\end{equation}
We note, though, that (to the best of our knowledge) the convergence proofs in \cite{zhao2018federated}  hold \textit{only} in the case of uniform client availability, i.e., $p_i = \frac{1}{N}$ for all $i$; in fact, in this case the above rule coincides with the agnostic FedAvg aggregation update (line $5$ of Algorithm \ref{alg:agnostic-fedavg}).

% \[
% \hat{\theta}^t_{\text{Weighted}} = \frac{N}{M} \sum_{i \in S^t} \frac{1}{N} \theta_i^{t-1} = \frac{1}{M} \sum_{i \in S^t} \theta_i^{t-1} = \hat{\theta}^t_{\text{Agnostic}}.
% \]

We conduct two experiments: one on the MNIST classification dataset, and another on a synthetic convex linear regression task. In both cases, data is partitioned across $N = 100$ users. At each global round, a fixed number of $M = 10$ users is selected to participate in a skewed, non-uniform manner: The skewness is introduced by sampling $M$ out of $N$ participating users \textit{without replacement} from a fixed, exponentially biased prior distribution over user indices, induced by weights proportional to $\exp{(-i/10)}$.  This process is consistent with our survival model of Section \ref{sec:problem-setup} and results in marginal participation probabilities $p_i$ that are significantly skewed; their values are empirically estimated via repeated draws (for verification purposes in our simulations).

\begin{figure}[!t]
    \centering
    \includegraphics[width=0.495\linewidth]{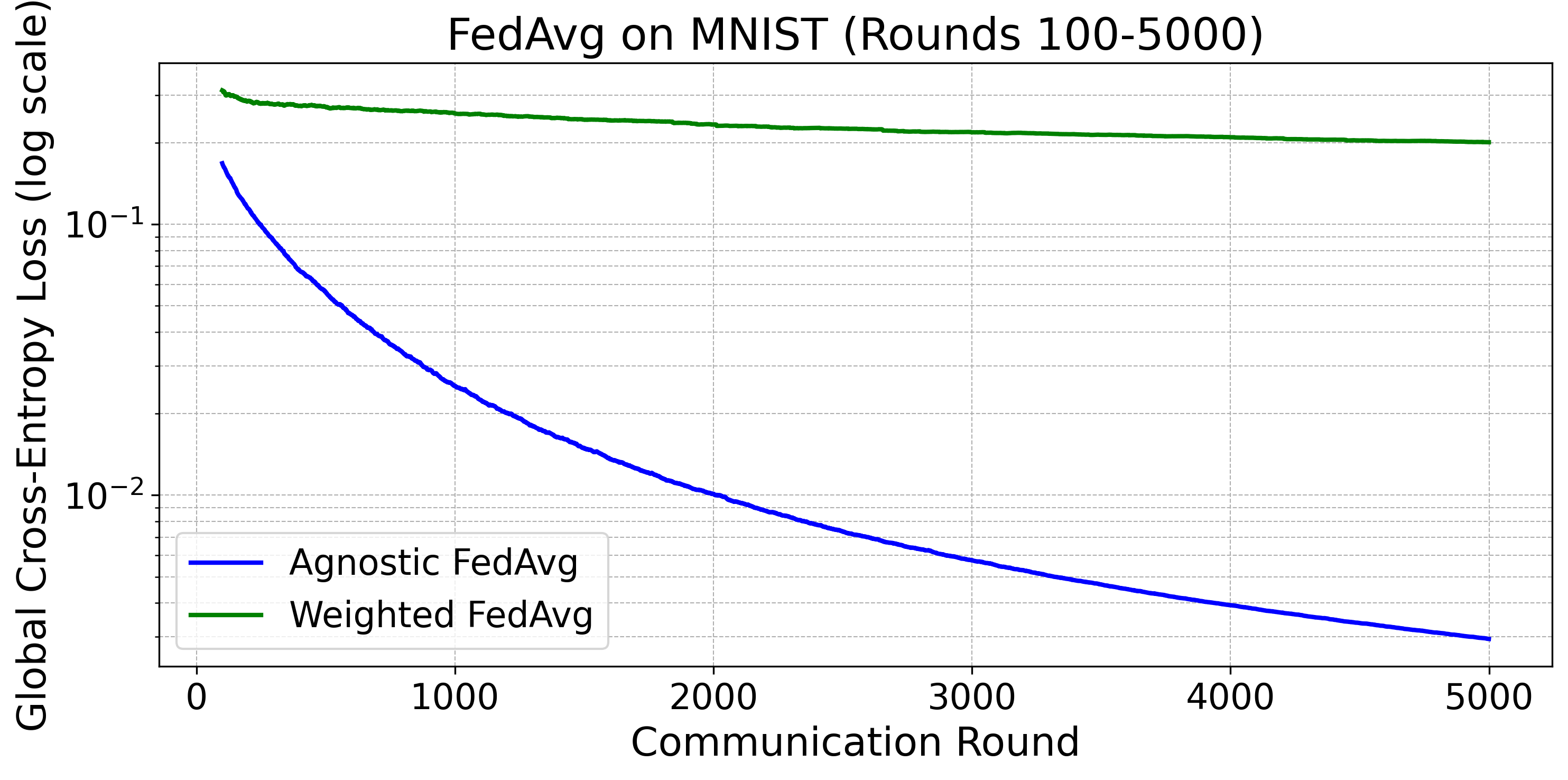}
    \hfill
    \includegraphics[width=0.495\linewidth]{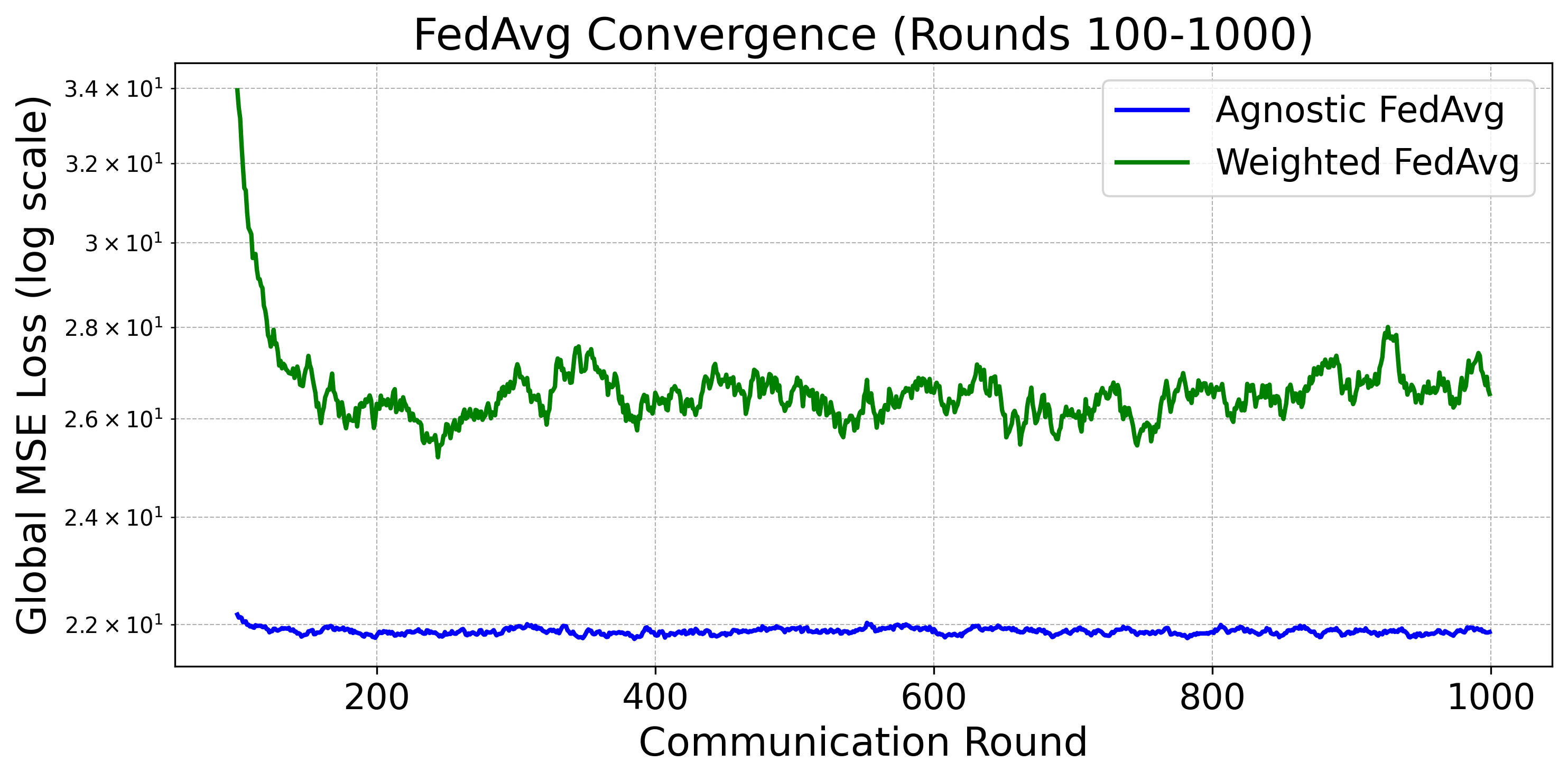}
    \vspace{-14pt}
    \caption{Left: FL on MNIST (log-scale cross-entropy loss). Right: FL on Linear Regression (log-scale MSE loss). Each curve represents average performance across five different random seeds.}
    \label{fig:fedavg_log_compare}
    \vspace{-5pt}
\end{figure}

As shown in Fig.~\ref{fig:fedavg_log_compare}, agnostic FedAvg (blue) consistently outperforms weighted FedAvg (green), despite the latter leveraging knowledge of the user selection probabilities $\{p_i\}_i$. This is particularly the case in the linear regression task, where the loss function is convex (and covered exactly by our analysis). These results suggest that, even when the $p_i$'s are known, the design of an aggregation policy that leverages this information (i.e., the $p_i$'s) remains an open and nontrivial question.
It is important to emphasize that agnostic FedAvg does not(operationally) rely on assumptions on \textit{the number} of available users at each round; this number can be random at each round, as also reflected in our convergence analysis. In contrast, prior literature assumes either full participation or \textit{fixed-size} client sampling. Even under fixed-size participation, agnostic FedAvg achieves the best performance without using $p_i$ as illustrated in Fig.~\ref{fig:fedavg_log_compare}, again in agreement with our analysis.

\begin{figure}[!t]
    %\centering
    \hspace{-4pt}\includegraphics[width=1.04\linewidth]{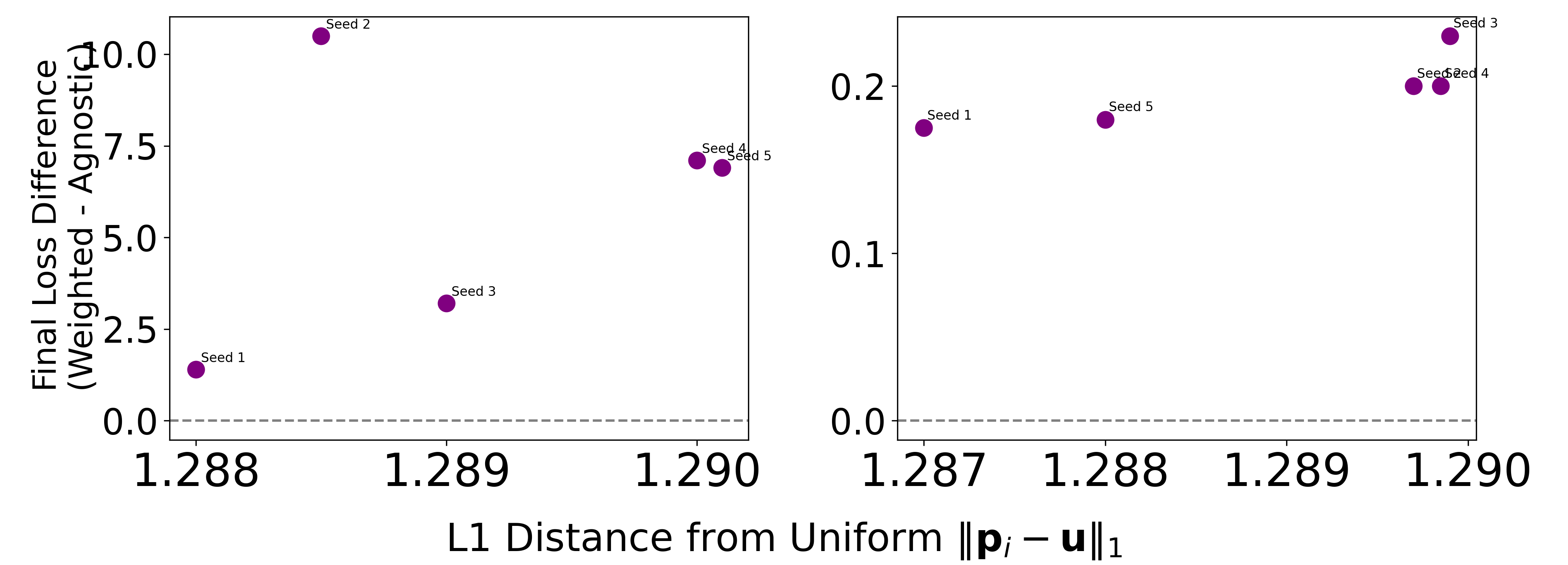}
    \vspace{-16pt}
    \caption{Loss difference (``weighted -- agnostic") versus participation skew ($\|\mathbf{p} - (1/N)\mathbf{1}\|_1$) across five random seeds for  MNIST classification (left) and linear regression (right). Each dot denotes a seed.}
    \label{fig:loss_vs_skew}
    \vspace{-16pt}
\end{figure}

To further analyze the discrepacy between the two schemes, Fig.~\ref{fig:loss_vs_skew} shows the (converged) \textit{signed} loss difference between weighted and agnostic FedAvg against the participation skew $\|\mathbf{p} -(1/N)\mathbf{1}\|_1$ (the total variation metric relative to the uniform distribution). In both tasks, the performance degradation of weighted FedAvg correlates favorably with increasing participation skew, implying that imbalanced client availability adversely affects effectiveness as compared with agnostic FedAvg. For implementation details, including how the participation skews were generated, please refer to the accompanying \href{https://github.com/HerlockSholmesm/Convergence-of-Agnostic-Federated-Averaging.git}{GitHub} repository.

\vspace{-8pt}
\section{Conclusion}
\vspace{-8pt}
In this work, we analyzed the convergence behavior of FedAvg under general, unknown, random and possibly non-uniform device participation %and established that the widely used Weighted FedAvg aggregation rule fails to converge in general when user availability is imbalanced. In contrast, 
We rigorously showed that \emph{agnostic FedAvg} optimizes a well-defined objective governed by user availability (not arbitrary preferences) and achieves convergence under this objective for convex nonsmooth loss function, with a standard rate of $\mathcal{O}(1/\sqrt{T})$. 

Our results bridge a significant gap in the literature by generalizing the convergence theory of (agnostic) FedAvg beyond the common but limiting assumptions of full-device participation or uniform fixed-size sampling. Unlike FL schemes that require knowledge of client probabilities $\{p_i\}_{i}$, agnostic FedAvg makes no such assumptions, yet remains provably convergent and empirically effective.

Through experiments, we also demonstrated that agnostic FedAvg consistently outperforms traditional weighted FedAvg, i.e., even in the fixed-size partial participation setting where the $p_i$'s are known and utilized by the server. This raises a fundamental open question: Can we design improved aggregation policies (at least empirically justifiable) that leverage potential knowledge of the $p_i$'s while still provably ensuring convergence? We pose this question as a potentially interesting topic for future reseach.
\vfill\pagebreak

% \section{REFERENCES}
% \label{sec:refs}
\clearpage

\bibliographystyle{IEEEtran}
\bibliography{refs}
\end{document}